# Reduction of Computational Complexity in Bayesian Networks through Removal of Weak Dependences

**Uffe Kjærulff**
Department of Mathematics and Computer Science, Aalborg University
Fredrik Bajers Vej 7E, DK-9220 Aalborg Ø, Denmark
uk@iesd.auc.dk

## Abstract

The paper presents a method for reducing the computational complexity of Bayesian networks through identification and removal of weak dependences (removal of links from the (moralized) independence graph). The removal of a small number of links may reduce the computational complexity dramatically, since several fill-ins and moral links may be rendered superfluous by the removal. The method is described in terms of impact on the independence graph, the junction tree, and the potential functions associated with these. An empirical evaluation of the method using large real-world networks demonstrates the applicability of the method. Further, the method, which has been implemented in Hugin, complements the approximation method suggested by Jensen & Andersen (1990).

## 1  INTRODUCTION

Decision making in domains with inherent uncertainty using Bayesian (belief) networks and exact computations often involve very high dimensional probability tables. Hence, for many practical problems, exact computations are prohibitive. Therefore, approximate solutions are often the best that can be hoped for. Such solutions can be provided through simulation or model simplification. We shall address the latter, although methods of the former type shall play an important role in our approach, which involves enforcement of additional conditional independence assumptions through removal of links from the moralized independence graph.

To illustrate the approach, consider the following toy example. Assume that dyspnoea (shortness of breath) ($d$) can be caused by one or more of the 'diseases' coughing ($c$), bronchitis ($b$), and lung cancer ($l$), and further that bronchitis causes coughing (Figure 1a). This model suggests marginal independence between bronchitis and lung cancer (shorthand: $b \perp\!\!\!\perp l$), and between coughing and lung cancer ($c \perp\!\!\!\perp l$). It might, however, be quite sensible to replace $c \perp\!\!\!\perp l$ with $c \perp\!\!\!\perp l \,|\, (b,d)$; that is, conditional independence between coughing and lung cancer given bronchitis and dyspnoea. The independence graph of this alternative model could be achieved through replacement of the directed link from coughing to dyspnoea with an undirected one, whereby the chain graph of Figure 1b emerges. Semantically, this implies that the relationship between coughing and dyspnoea is non-causal. (Note that an independence graph equivalent to that of Figure 1b might be obtained by simple reversal of the directed link from coughing to dyspnoea.)

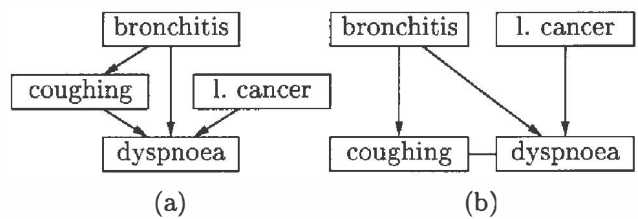

Figure 1: Removal of the moral link between coughing and lung cancer in part (a) results in the 'less demanding' independence graph of part (b).

Specification of conditional probabilities for model (a) involves a four-dimensional table for $(d \,|\, b, c, l)$ and a two-dimensional one for $(c \,|\, b)$, whereas for model (b) it suffices to specify two three-dimensional tables, one for $(c, d \,|\, b)$ and one for $(d \,|\, b, l)$. If, for example, each of the four variables is described in terms of five discrete states, this means that model (a) requires specification of $4 \cdot 5^3 + 4 \cdot 5 = 520$ conditional probabilities, whereas model (b) requires 'only' $24 \cdot 5 + 4 \cdot 5^2 = 220$.

Using the suggested approximation method, model (b) can be obtained from model (a) by removal of the moral link between coughing and lung cancer.

Briefly, the method provides a systematic way of performing model transformations as illustrated in Figure 1 such that one additional conditional independence assumption is explicitly being enforced (and possibly some implicit ones, which follow naturally) and such that an (sub)optimal balance between reduction of computational complexity and approximation error is achieved. A candidate class (explicit) assumption takes the form $\alpha \perp\!\!\!\perp \beta \,|\, C \setminus \{\alpha, \beta\}$, where $C$ is a clique in a junction tree corresponding to an independence graph $\mathcal{G}$, such that $\alpha$ and $\beta$ are connected in the moral graph corresponding to $\mathcal{G}$ and such that $C$ is the unique clique containing both $\alpha$ and $\beta$. That is, the method aims at splitting (large) cliques into smaller ones while keeping a small 'distance' between the exact and the approximate distributions. This distance is computed using either exact or simulated clique potentials of a (imaginary) junction tree, where the storage requirements of simulated potentials, obtained through Monte-Carlo sampling, depends only linearly on both the clique size and the sample size.

The rest of the paper is organized as follows. Section 2 reviews the key features of graphical chain models and junction trees necessary for the presentation. Section 3 presents the method, including descriptions of its impact on the junction tree, the independence graph, and the potential functions associated with these. Please note that the results are stated without proofs; the interested reader is referred to Kjærulff (1993). Section 4 demonstrates the applicability of the method by presenting some results of applying it on large real-world networks. Section 5 summarizes the features of the presented approach and argues that it complements the approach of Jensen & Andersen (1990).

For a discussion of the choice of criterion for selecting the optimal link to remove and a presentation of the implications of link removal in terms of correctness of inference, the reader is referred to Kjærulff (1993).

## 2   GRAPHICAL CHAIN MODELS AND JUNCTION TREES

The term Bayesian networks has traditionally been used as a synonym for recursive graphical models (Wermuth & Lauritzen 1983) for which the independence structure is encoded by directed acyclic graphs. In the present paper we shall, however, use 'Bayesian networks' as a synonym for the more general class of models denoted graphical chain models (Lauritzen & Wermuth 1984, Lauritzen & Wermuth 1989) for which the independence structure is encoded by chain graphs. Notice that the class of graphical chain models also contains the subclass of graphical models (Darroch, Lauritzen & Speed 1980) with independence structure encoded by undirected graphs.

### 2.1   CHAIN GRAPHS

In the following the notion of chain graphs shall be reviewed briefly and fairly informally. For a more thorough treatment of the subject see e.g. Frydenberg (1989).

Let $\mathcal{G} = (V, E = E^d \cup E^u)$ be a graph with nodes (vertices) $V$ and links (edges) $E \subseteq V \times V$, where $E^d =$ $\{(\alpha, \beta) \in E \mid (\beta, \alpha) \notin E\}$ is the subset of *directed links* and $E^u = \{(\alpha, \beta) \in E \mid (\beta, \alpha) \in E\}$ the subset of *undirected links*. If there is a link between $\alpha$ and $\beta$, denoted $\alpha \sim \beta$, they are said to be *connected*. A directed link between $\alpha$ and $\beta$ is denoted $\alpha \to \beta$ or $\alpha \leftarrow \beta$, and an undirected link is denoted $\alpha \,\text{---}\, \beta$. We shall use $\alpha \sim \beta$, etc. to denote either '$\alpha$ and $\beta$ are connected' or 'the link between $\alpha$ and $\beta$' depending on the context.

A *path* $\langle \alpha = \alpha_1, \ldots, \alpha_k = \beta \rangle$ from $\alpha$ to $\beta$ in $\mathcal{G}$ is an ordered sequence of distinct nodes such that $\alpha_i \sim \alpha_{i+1}$ for each $i = 1, \ldots, k-1$. The path is *undirected* if $\alpha \,\text{---}\, \beta$ for each $i = 1, \ldots, k-1$. The path is *directed* if either $\alpha \,\text{---}\, \beta$ or $\alpha_i \to \alpha_{i+1}$ for each $i = 1, \ldots, k-1$ and the path includes at least one directed link. A *cycle* is a path $\pi = \langle \alpha = \alpha_1, \ldots, \alpha_k = \beta \rangle$ with the exception that $\alpha = \beta$.

For $A, B, C \subseteq V$, $C$ is said to *separate* $A$ from $B$ if for all paths $\langle \alpha = \alpha_1, \ldots, \alpha_k = \beta \rangle$, where $\alpha \in A$ and $\beta \in B$, $\{\alpha_1, \ldots, \alpha_k\} \cap C \neq \emptyset$. A graph $\mathcal{G}$ is *connected* if there is a path between each pair of nodes of $\mathcal{G}$. Unless otherwise stated, connectivity shall henceforth be assumed.

Now, $\mathcal{G}$ is a *chain graph* if it contains no directed cycles. If $\mathcal{G}$ is a chain graph, then $\{K_1, \ldots, K_n\}$ are called the *chain components* of $\mathcal{G}$ if $\{K_1, \ldots, K_n\}$ is the set of connected components of $(V, E^u)$. There are two important special classes of chain graphs. If $n = |V|$ (i.e., one node per chain component), $\mathcal{G}$ is called a *directed acyclic graph* (DAG). If $n = 1$, $\mathcal{G}$ is called an *undirected graph*.

A subset $A \subseteq V$ induces a *subgraph* $\mathcal{G}_A = (A, E_A)$ of $\mathcal{G}$, where $E_A = E \cap (A \times A)$. (Note that any subgraph of a chain graph is a chain graph.) A graph is *complete* if all nodes are pairwise connected. A subset $A \subseteq V$ is complete if it induces a complete subgraph, and if $A$ is maximal (i.e., there is no complete subset $B \subseteq V$ such that $A \subset B$), then it is called a *clique*.

The *parents* of $A \subseteq V$ is the subset $\text{pa}(A) \subseteq V \setminus A$ such that for each $\beta \in \text{pa}(A)$ there is an $\alpha \in A$ for which $\beta \to \alpha$. The set of *children* of $A$, denoted $\text{ch}(A)$, is defined analogously. The *neighbours* of $A$ is the subset $\text{nb}(A) \subseteq V \setminus A$ such that for each $\beta \in \text{nb}(A)$ there is an $\alpha \in A$ for which $\alpha \,\text{---}\, \beta$. The *ancestral set* of $A \subseteq V$ is the subset $\text{An}(A) \subseteq V$ such that for each $\beta \in \text{An}(A)$ either $\beta \in A$ or there is a directed or undirected path from $\beta$ to at least one $\alpha \in A$.

The moral graph $\mathcal{G}^m$ of a chain graph $\mathcal{G}$ is obtained by first adding undirected links between each pair of unconnected nodes in $\text{pa}(K)$ for each chain component $K$, and then replacing all directed links by undirected ones.

An undirected graph $\mathcal{G} = (V, E)$ is *triangulated* (also: *decomposable* or *chordal*) if each cycle of length greater than 3 has a chord (i.e., a link between two non-consecutive nodes of the cycle).

## 2.2  GRAPHICAL CHAIN MODELS

For a chain graph $\mathcal{G} = (V, E)$ we consider a collection of discrete random variables $(X_\alpha)_{\alpha \in V}$ taking values in probability spaces $\mathrm{Sp}(X_\alpha)$. For brevity we shall interchangeably refer to $\alpha \in V$ as both a node and a variable. Thus we shall write e.g. $\alpha$ instead of $X_\alpha$. For a subset $A \subseteq V$ we let $\mathrm{Sp}(A) = \times_{\alpha \in A} \mathrm{Sp}(\alpha)$ (i.e., the Cartesian product of the state spaces of the variables in $A$).

A probability function $p = p_V$ is said to *factorize* according to a chain graph $\mathcal{G} = (V, E)$ if there exist non-negative functions $\phi_A$ defined on $\mathrm{Sp}(A)$ such that

$$p \propto \prod_{A \in \mathcal{A}} \phi_A, \qquad (1)$$

where $\mathcal{A}$ is the set of cliques of $\mathcal{G}^m$. The functions $\phi_A$ shall be called *component potentials* of $p$. For $\mathcal{G}$ being a DAG this simplifies to

$$p = \prod_{v \in V} p(v \mid \mathrm{pa}(v)). \qquad (2)$$

A similar factorization exists in the general case. Let namely

$$p(K \mid \mathrm{pa}(K)) = \left( \prod_{A \in \mathcal{A}_K} \phi_A \right) \Big/ \left( \sum_K \prod_{A \in \mathcal{A}_K} \phi_A \right) \qquad (3)$$

where $K$ is a chain component of $\mathcal{G}$ and $\mathcal{A}_K = \{A \in \mathcal{A} \mid A \subseteq K \cup \mathrm{pa}(K), A \cap K \neq \emptyset\}$. Then

$$p = \prod_{K \in \mathcal{K}} p(K \mid \mathrm{pa}(K)), \qquad (4)$$

where $\mathcal{K}$ is the set of chain components of $\mathcal{G}$.

If $p$ factorizes according to $\mathcal{G}$, then $\mathcal{G}$ is said to be an *independence graph* of $p$, and $p$ is a graphical chain model (a probability function of a Bayesian network with $\mathcal{G}$ as underlying graph). (The phrase '$p$ is Markov with respect to $\mathcal{G}$' is a synonym for '$p$ factorizes according to $\mathcal{G}$'.)

In the special case of $\mathcal{G} = (V, E)$ being a DAG all conditional independence statements captured by $\mathcal{G}$ can be found using the d-separation criterion of Pearl (1988) or the equivalent criterion of Lauritzen, Dawid, Larsen & Leimer (1990). But in the general case the Markov properties (i.e., conditional independence properties) captured by $\mathcal{G}$ are expressed by the following theorem (Frydenberg 1989).

**Theorem 1** *Let $p$ factorize according to a chain graph, $\mathcal{G} = (V, E)$. Then $A \perp\!\!\!\perp B \mid C$ with respect to $p$ for any subsets $A, B, C \subseteq V$ whenever $C$ separates $A$ from $B$ in $(\mathcal{G}_{\mathrm{An}(A \cup B \cup C)})^m$.*

Note that the formulation of this theorem, describing the *global chain Markov property*, is identical to the theorem of Lauritzen et al. (1990) describing the *directed global Markov property* for recursive graphical models (i.e., where $\mathcal{G}$ is a DAG).

## 2.3  JUNCTION TREES

By exploiting the conditional independence relations among the variables of a Bayesian network, the underlying joint probability space may be decomposed into a set of subspaces corresponding to a decomposable (hypergraph) cover of the moralized graph such that exact inference can be performed by simple message passing in a maximal spanning tree of the cover (Lauritzen & Spiegelhalter 1988, Jensen 1988, Jensen, Lauritzen & Olesen 1990). Technically, a decomposable cover of a Bayesian network with underlying chain graph $\mathcal{G}$ is created by triangulating $\mathcal{G}^m$ (i.e., adding undirected links, so-called *fill-ins*, to $\mathcal{G}^m$ to make it triangulated). That is, the set of cliques of the triangulated graph is a decomposable cover of the network.

Jensen (1988) has shown that any maximal spanning tree of a decomposable cover, $\mathcal{C}$, can be used as the basis for a simple inward/outward message-passing scheme for propagation of evidence (belief updating) in Bayesian networks, where maximality is defined in terms of the sum of cardinalities of the intersections between adjacent nodes (cliques) of the tree. Jensen named these trees *junction trees*. The intersections between neighbouring cliques of a junction tree are called *separators* (Jensen et al. 1990).

We shall henceforth refer to a junction tree by the pair $(\mathcal{C}, \mathcal{S})$ of cliques and separators. It can be shown that for each path $(C = C_1, \ldots, C_k = D)$ in a junction tree, $C \cap D \subseteq C_i$ for all $1 \leq i \leq k$, implying that $A \perp\!\!\!\perp B \mid S$ for each $S \in \mathcal{S}$, where $A$ and $B$ are the sets of variables of the two subtrees (except $S$) induced by the removal of the link corresponding to $S$ (Jensen 1988).

To each clique and each separator is associated a *belief potential*, $\phi_A$. The joint probability distribution, $p_V$, of a Bayesian network with a junction tree $(\mathcal{C}, \mathcal{S})$ is proportional to the *joint (system) belief* $\phi_V$ given by

$$p_V \propto \phi_V = \frac{\prod_{C \in \mathcal{C}} \phi_C}{\prod_{S \in \mathcal{S}} \phi_S}. \qquad (5)$$

A belief potential $\phi_A$ is *normalized* if $\sum_A \phi_A = 1$. If all belief potentials of a junction tree are normalized, then $\phi_V$ is normalized (i.e., $p_V = \phi_V$).

A junction tree $\Upsilon = (\mathcal{C}, \mathcal{S})$ is said to be *consistent* if

$$\sum_{C \setminus D} \phi_C \propto \sum_{D \setminus C} \phi_D \quad \text{for all } C, D \in \mathcal{C}$$

(i.e., the marginal potentials for $C \cap D$ with respect to $\phi_C$ and $\phi_D$ are proportional). Consistency of $\Upsilon$ shall interchangeably be referred to as consistency of its associated joint belief, $\phi_V$.

## 3  ENFORCING INDEPENDENCE ASSUMPTIONS

The computational complexity imposed by a particular junction tree $(\mathcal{C}, \mathcal{S})$ is roughly determined by the clique, $C \in \mathcal{C}$, with the largest state space. Thus by splitting $C$ into smaller cliques a significant reduction of the computational complexity might be obtained.

If $\{\alpha, \beta\} \subseteq C$ such that there is no other clique in $\mathcal{C}$ containing $\{\alpha, \beta\}$, then adding $\alpha \perp\!\!\!\perp \beta \,|\, C \setminus \{\alpha, \beta\}$ to the set of independence statements amounts to splitting $C$ into $C_\alpha = C \setminus \{\beta\}$ and $C_\beta = C \setminus \{\alpha\}$, which might or might not become new cliques of the modified junction tree (see the examples of Figure 2).

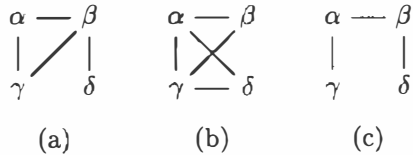

Figure 2: Removal of $\alpha - \beta$: (a): both $\{\alpha, \gamma\}$ and $\{\beta, \gamma\}$ become new cliques; (b): $\{\beta, \gamma\}$ become a new clique, but $\{\alpha, \gamma\}$ does not; (c): neither $\{\alpha\}$ nor $\{\beta\}$ become new cliques.

The requirement that $C$ must be the only clique containing $\{\alpha, \beta\}$ ensures that $\alpha \perp\!\!\!\perp \beta \,|\, C \setminus \{\alpha, \beta\}$ or, equivalently, that the graph obtained by removing $\alpha - \beta$ in the triangulated graph corresponding to $(\mathcal{C}, \mathcal{S})$ is triangulated; see Kjærulff (1993) for details.

## 3.1   AN EXAMPLE

To understand the main issues of the proposed approximation method we shall present a small example. Consider the sample chain graph of Figure 3a with corresponding moral graph of Figure 3b (solid links). The dashed link is a fill-in added to make the graph triangulated. The junction tree corresponding to the triangulated graph of Figure 3b is shown in Figure 4a.

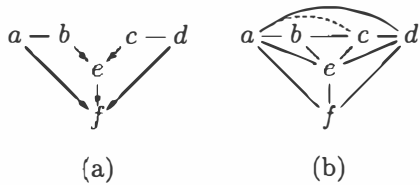

(a)                           (b)

Figure 3: (a) Sample independence graph. (b) Corresponding moral graph (solid links) and a triangulated graph (all links).

Reduction of the computational complexity of the junction tree could be accomplished by extending the set of conditional independence statements displayed by the tree. Adding e.g. the statement $c \perp\!\!\!\perp d \,|\, (a, e)$ (i.e., removal of $c - d$ from the triangulated graph) causes clique $\{a, c, d, e\}$ to split into the sets $\{a, d, e\}$ and $\{a, c, e\}$ neither of which appear to be cliques of the reduced graph (Figure 4b).

Since we wish to add just one statement to the set $\mathcal{I}$ of independence statements displayed by the original independence graph of Figure 3a, the revised independence graph is, in general, not obtained through simple link removal. Removal of $c - d$ in Figure 3a would induce several new independence statements ($c \perp\!\!\!\perp d$, $d \perp\!\!\!\perp e$, $a \perp\!\!\!\perp d \,|\, e$, etc.) which do not follow as natural consequences of $c \perp\!\!\!\perp d \,|\, (a, e)$. The set of independence statements displayed by each chain graph of Figure 5 is a subset of $\mathcal{I}^* = \mathcal{I} \cup \{c \perp\!\!\!\perp d \,|\, (a, e)\}$. This follows from the fact that the three moral graphs are identical to the moral graph of Figure 3b with $c - d$ removed. Thus, each graph of Figure 5 is a *correct* representation of $\mathcal{I}^*$, but none of the graphs are *perfect* representations, since they fail to represent e.g. the statements $d \perp\!\!\!\perp e \,|\, c$ and $a \perp\!\!\!\perp e \,|\, b$.

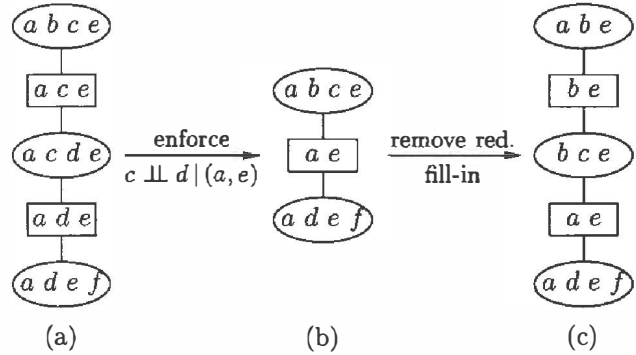

Figure 4: (a) Junction tree corresponding to Figure 3b. (b) Removal of $c - d$ causes clique $\{a, c, d, e\}$ to disappear. (c) The fill-in $a - c$ is rendered redundant, splitting clique $\{a, b, c, e\}$ into two smaller ones.

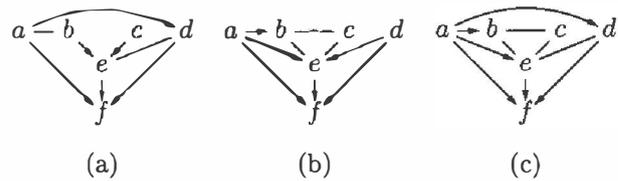

Figure 5: Competing independence graphs obtained by adding $c \perp\!\!\!\perp d \,|\, (a, e)$.

Notice that the moral graph corresponding to the graphs of Figure 5 is triangulated. This eliminates the need for the fill-in between $a$ and $c$, allowing clique $\{a, b, c, e\}$ to be split into the two smaller cliques $\{a, b, e\}$ and $\{b, c, e\}$ (Figure 4c). In general, a possibly large number of fill-ins and moral links might be rendered redundant by the removal of a single link. If, for example, $b \to e$ is removed in Figure 3a, the moral link $b - c$ disappears.

Enforcement of the conditional independence statement $c \perp\!\!\!\perp d \,|\, (a, e)$ thus provided a reduction of complexity in terms of sizes of cliques from three 4-cliques (i.e., cliques of four variables) to one 4-clique and two 3-cliques. This corresponds to at least a 37% reduction of space requirements (binary variables) even though the resulting independence graph(s) at first glance seems more 'complicated'.

## 3.2   OUTLINE OF METHOD

The above example provided insight into some of the issues related to the approximation method. Before presenting the technicalities of the method, let us summarize the underlying philosophy and list the issues to be dealt with in more detail.

When attempts to compile a Bayesian network into a junction tree fails on account of excessive memory requirements, the problems are often caused by a small number of cliques. The proposed method is based on the idea of splitting these cliques into smaller ones (i.e., extending the set of independence statements). Therefore, the first step is to create a junction tree with exact or simulated clique potentials. (Although exact clique potentials can be created, there might still be a wish to reduce the space requirements if this can be done without attaining an unacceptable level of imprecision.)

Clique potentials (whether exact or simulated) must be provided such that the deviation between these 'correct' potentials and the approximate ones can be computed. These measures of deviation (or distance) must then be used as the basis of a criterion for selecting the link to be removed.

Simulated clique potentials can be provided through various kinds of Monte-Carlo simulation like Gibbs sampling and 'forward sampling' which have complexities proportional to the moral graph. We shall not discuss this issue any further, even though there are some interesting points concerning optimal choice of simulation method, especially when the underlying independence graph is not a DAG.

A Bayesian network with underlying probability model $p$ may be exhaustively described in terms of four components: (1) a potential representation of $p$ based on component potentials (cf. Equation (1)), (2) an independence graph, $\mathcal{G}$, of $p$, (3) a junction tree (decomposable hypergraph cover of $\mathcal{G}^m$), and (4) a potential representation of $p$ based on belief potentials (cf. Equation (5)). Notice that it suffices to include one of the potential representations for an exhaustive description of a Bayesian network. We shall, however, include them both as a matter of convenience.

We shall now detail the impacts on these four components when removing $\alpha \sim \beta$ from the moral graph.

### 3.3 BELIEF POTENTIALS

Let $\Upsilon = (\mathcal{C}, \mathcal{S})$ be a junction tree, $C \in \mathcal{C}$ the unique clique containing $\{\alpha, \beta\}$, and $\phi$ a consistent joint belief for $\Upsilon$. Let further

$$\psi_C = \frac{\sum_\alpha \phi_C \sum_\beta \phi_C}{\sum_{\alpha, \beta} \phi_C} \quad \Leftrightarrow \quad \alpha \perp\!\!\!\perp \beta \mid C \setminus \{\alpha, \beta\}$$

with respect to $\psi$. Since

$$\sum_\gamma \phi_C = \sum_\gamma \psi_C, \quad \gamma \in \{\alpha, \beta\},$$

and $C$ is the unique clique containing $\{\alpha, \beta\}$ it follows that for each separator, $S$, between $C$ and its neighbours in $\Upsilon$ either $S \subseteq C \setminus \{\alpha\}$ or $S \subseteq C \setminus \{\beta\}$ implying

$\psi_S = \phi_S$. That is, the potentials of the (possible) new cliques are $\phi_{C \setminus \{\alpha\}}$ and $\phi_{C \setminus \{\beta\}}$, and the potentials of the cliques in $\mathcal{C} \setminus \{C\}$ remain unaltered.

### 3.4 JUNCTION TREE

Let $C_1, \ldots, C_k$ be the neighbours of $C$ in a junction tree $\Upsilon = (\mathcal{C}, \mathcal{S})$ and $S_1, \ldots, S_k$ the associated separators, where $C$ is the unique clique containing $\{\alpha, \beta\}$. As demonstrated in Figure 2, the removal of the link between $\alpha$ and $\beta$ produces two, one, or zero new cliques. That is, (a) both $C_\alpha = C \setminus \{\beta\}$ and $C_\beta = C \setminus \{\alpha\}$ are cliques in the revised junction tree $\Upsilon'$, (b) $C_\alpha$ ($C_\beta$) is a clique in $\Upsilon'$ and $C_\beta$ ($C_\alpha$) is not, or (c) neither $C_\alpha$ nor $C_\beta$ are cliques in $\Upsilon'$. It is easy to see that $\Upsilon'$ is constructed from $\Upsilon$ as indicated in Figure 6, where the dashed parts illustrate the cliques, separators, and links to be added to $\Upsilon$ (with $C$ and its incident links removed) and the dotted parts the separators and links to be removed; see Kjærulff (1993) for details.

Note that in all three cases we have $S = C \setminus \{\alpha, \beta\}$ meaning that $S$ separates $\Upsilon'$ into two subtrees $\Upsilon'_A = (\mathcal{C}'_A, \mathcal{S}'_A)$ and $\Upsilon'_B = (\mathcal{C}'_B, \mathcal{S}'_B)$, where $A$ and $B$ are the corresponding sets of variables such that $\alpha \in A$ and $\beta \in B$. From the discussion in Section 3.3 it follows trivially that

$$\psi_{A \cup S} = \frac{\prod_{C' \in \mathcal{C}'_A} \psi_{C'}}{\prod_{S' \in \mathcal{S}'_A} \psi_{S'}} = \frac{\prod_{C' \in \mathcal{C}'_A} \phi_{C'}}{\prod_{S' \in \mathcal{S}'_A} \phi_{S'}} = \phi_{A \cup S}$$

and similarly $\psi_{B \cup S} = \phi_{B \cup S}$. Therefore,

$$\psi = \frac{\phi_{A \cup S} \phi_{B \cup S}}{\phi_S}. \tag{6}$$

The reduction, $\sigma(\alpha, \beta)$, of the computational complexity achieved by the removal of $\alpha - \beta$ can be expressed as (a) $\|C\| - (\|C_\alpha\| + \|C_\beta\| + \|S\|)$, (b) $\|C\| - (\|C_\alpha\| + \|S\|) + \|S_k\|$, or (c) $\|C\| - \|S\| + \|S_1\| + \|S_k\|$, where $\| \cdot \| = |Sp(\cdot)|$; cf. Figure 6. This can be expressed compactly as

$$\sigma(\alpha, \beta) = \|C\| \left(1 - I_\alpha \|\beta\|^{-1} - I_\beta \|\alpha\|^{-1}\right) - \|S\| + (1 - I_\alpha)\|S_1\| + (1 - I_\beta)\|S_k\|, \tag{7}$$

where $I_\gamma = 1$ ($\gamma \in \{\alpha, \beta\}$) if $C_\gamma$ is a clique and 0 otherwise. Note that $-\|S\| \leq \sigma \leq \|C\| - \|S\| + \|S_1\| + \|S_k\|$, where $\sigma$ reaches its lower bound when $\|\alpha\| = \|\beta\| = 2$ and $I_\alpha = I_\beta = 1$, and its upper bound when $I_\alpha = I_\beta = 0$.

### 3.5 INDEPENDENCE GRAPH

Since $\psi_A = \psi_{A \cup S} = \phi_{A \cup S}$, the independence relations among the variables of the set $A$ remain unaltered by the removal of $\alpha - \beta$, where $A$, $B$, $S$, $\phi$, and $\psi$ are given in Section 3.4. The same applies to $B$. That is, the marginal independence graphs for $A$, $B$ and $S$ are identical for $\phi$ and $\psi$. Therefore, the problem of determining the independence graph of $\psi$ may be formulated as the problem of combining marginal independence graphs such that the independence statements expressed by these are not violated and such that the combined graph represents the fact that $A \perp\!\!\!\perp B \mid S$ (or $A \setminus S \perp\!\!\!\perp B \setminus S \mid S$ to be exact).

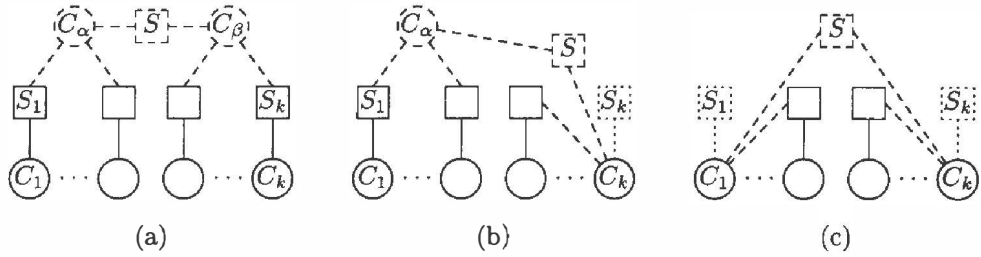

Figure 6: Removal of the link between $\alpha$ and $\beta$ results in a junction tree with a new separator $S = C \setminus \{\alpha, \beta\}$ separating the tree into a subtree containing $\alpha$ but not $\beta$ and a subtree containing $\beta$ but not $\alpha$. In parts (b) and (c) we assume, without loss of generality, that $C_\alpha \subset C_1$ (part (c) only) and $C_\beta \subset C_k$ (i.e., $C_\alpha = S_1$ and $C_\beta = S_k$).

Given an independence graph of a probability function (belief potential), $p = p_V$, the following theorem provides a way of establishing an independence graph of any marginal $p_A$, $A \subseteq V$.

**Theorem 2** *Let the chain graph $\mathcal{G} = (V, E)$ be an independence graph of $p = p_V$ and $\alpha \in V$. Then $\mathcal{G}^{\downarrow}_{V \setminus \{\alpha\}} = (V \setminus \{\alpha\}, E^{\downarrow}_{V \setminus \{\alpha\}})$ is an independence graph of $p_{V \setminus \{\alpha\}} = \sum_\alpha p_V$, where $\mathcal{G}^{\downarrow}_{V \setminus \{\alpha\}}$ is constructed from $\mathcal{G}$ by rendering $nb(\alpha)$ complete by adding undirected links if necessary, adding $\beta \to \gamma$ for each $\beta \in pa(\alpha)$ and $\gamma \in nb(\alpha) \cup ch(\alpha)$, unless $\beta \sim \gamma$, adding $\beta \to \gamma$ for each $\beta \in nb(\alpha)$ and $\gamma \in ch(\alpha)$, unless $\beta \sim \gamma$, rendering $ch(\alpha)$ complete in such a way that no directed paths are introduced, and removing $\alpha$ and the links incident to it.*

In proving Theorem 2, it is profitable to note that correctness of $\mathcal{G}' = \mathcal{G}^{\downarrow}_{V \setminus \{\alpha\}}$ follows if separation of $A$ and $B$ by $C$ in $(\mathcal{G}'_{\mathrm{An}(A \cup B \cup C)})^m$ implies separation in $(\mathcal{G}_{\mathrm{An}(A \cup B \cup C)})^m$ as well, and that perfectness of $\mathcal{G}^{\downarrow}_{V \setminus \{\alpha\}}$ follows if separation in $(\mathcal{G}_{\mathrm{An}(A \cup B \cup C)})^m$ implies separation in $(\mathcal{G}'_{\mathrm{An}(A \cup B \cup C)})^m$ provided $\mathcal{G}$ is perfect.

It should be noticed that perfectness of $\mathcal{G}$ does not necessarily imply perfectness of $\mathcal{G}^{\downarrow}_{V \setminus \{\alpha\}}$. The following example illustrates this point. Let $V = \{\alpha, \beta, \gamma, \delta, \varepsilon\}$ and let the DAG of Figure 7a be an independence graph of $p$. Since $\beta \not\!\perp\!\!\!\perp \delta \mid \{\gamma, \varepsilon\}$ with respect to $p$ (and $p_{V \setminus \{\alpha\}}$), $\beta$ and $\delta$ must be connected in an independence graph of $p_{V \setminus \{\alpha\}} = \sum_\alpha p$, and, since $\gamma \perp\!\!\!\perp \varepsilon$ and $\gamma \not\!\perp\!\!\!\perp \varepsilon \mid \beta$, a candidate independence graph of $p_{V \setminus \{\alpha\}}$ could be the one of Figure 7b. However, since $\gamma \perp\!\!\!\perp \varepsilon \mid \delta$ with respect to $p$ (and $p_{V \setminus \{\alpha\}}$), this graph is not perfect, but it is correct, since it does not represent non-existing independence statements. Thus, all the independence properties of $p_{V \setminus \{\alpha\}}$ cannot be represented by a single chain graph. If we want a perfect representation, a more sophisticated language must be adopted. One such language may be given by the class of annotated graphs (Geva & Paz 1992). However, in the present paper we shall refrain from pursuing this any further.

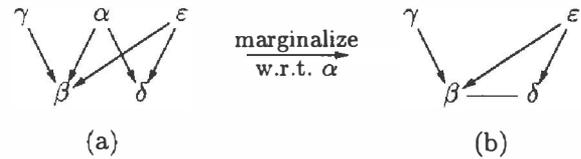

(a)        (b)

Figure 7: $\gamma \perp\!\!\!\perp \varepsilon \mid \delta$ with respect to to $p$ (and $\sum_\alpha p$) which is Markov with respect to the DAG in part (a). However, $\sum_\alpha p$ is not Markov with respect to the graph in part (b), since according to that $\gamma \not\!\perp\!\!\!\perp \varepsilon \mid \delta$.

Theorem 2 provides a method for constructing an independence graph of the marginal distribution $p_{V \setminus \{\alpha\}}$. However, the construction of an independence graph of the approximate joint belief $\psi = \phi_{A \cup S} \phi_{B \mid S}$ involves combination of a marginal independence graph and a conditional (marginal) independence graph.[1] The independence graph of the conditional distribution $p_{V \mid \alpha}$ is obtained simply by moralizing the subgraph induced by $\mathrm{An}(\alpha)$ and removing $\alpha$ and the links incident to it.

**Theorem 3** *Let the chain graph $\mathcal{G} = (V, E)$ be an independence graph of $p = p_V$ and let $E^m_{\mathrm{An}(A)}$ be the links of $(\mathcal{G}_{\mathrm{An}(A)})^m$. Then $\mathcal{G}' = (V, E \cup E^m_{\mathrm{An}(A)})$ is a chain graph and a conditional independence graph of $p_{V \mid A}$.*

By the methods of Theorem 2 and Theorem 3 we can construct any marginal independence graph (possibly conditional on a set of variables) by successive removal of the relevant variables.

Note that the presence of the set $A$ and the links incident to $A$ in the independence graph of $p_{V \mid A}$ is unnecessary for a correct interpretation of the conditional independence relations among variables in $V \setminus A$ given $A$. However, when combining a conditional and a marginal independence graph to obtain a joint independence graph, $A$ and some links connecting $A$ to $V \setminus A$ are needed. In fact, when constructing a conditional independence graph we shall proceed as follows.

---

[1] For brevity we shall refer to an independence graph of a marginal distribution as a marginal independence graph, and similarly for the conditional case.

**Corollary 1** *Let $p$ and $\mathcal{G}'$ be given as in Theorem 3. The graph obtained by (i) removing all links between nodes in $S$ and (ii) making all links between $S$ and $nb(S)$ undirected is a conditional independence graph of $p_{V \mid A}$.*

Theorem 4 below states that a joint independence graph can be formed by simple graph union of a conditional and a marginal independence graph.

**Theorem 4** *Let the chain graph $\mathcal{G}^{\downarrow}_{A \cup S} = (A, E^{\star}_A)$ be a marginal independence graph of $p_{A \cup S}$ and the chain graph $\mathcal{G}^{\downarrow}_{B \mid S} = (B, E^{\downarrow}_B)$ a conditional independence graph of $p_{B \mid S}$ complying with Corollary 1, where $A \cup B \cup S = V$ such that $A \cap B = S$ and $A \perp\!\!\!\perp B \mid S$ with respect to $p = p_V$. If $\mathcal{G}' = \mathcal{G}^{\downarrow}_{A \cup S} \cup \mathcal{G}^{\downarrow}_{B \mid S}$ is not a chain graph (i.e., it contains directed cycle(s)), replace links $\gamma - \delta$ with $\gamma \to \delta$, where $\gamma \in S$ and $\delta \in nb(S) \cap B$, until $\mathcal{G}'$ becomes a chain graph. Then $\mathcal{G}'$ is an independence graph of $p$. Further, $\mathcal{G}'$ is perfect if both $\mathcal{G}^{\downarrow}_{A \cup S}$ and $\mathcal{G}^{\downarrow}_{B \mid S}$ are perfect.*

Returning to the example in Section 3.1, we identify the sets $A = \{a, b, c, e\}$, $B = \{a, d, e, f\}$, and $S = \{a, e\}$. Following the above results we determine a marginal independence graph and a conditional one, and then combine these into a new joint independence graph. This combination can involve one of three principally different sets of marginal and conditional graphs: (1) marginal graph for $A \cup S$ plus conditional one for $B \mid S$, (2) marginal graph for $B \cup S$ plus conditional one for $A \mid S$, or (3) conditional graphs for $A \mid S$ and $B \mid S$ plus marginal one for $S$, reflecting the factorizations $\psi = \phi_{A \cup S} \phi_{B \mid S}$, $\psi = \phi_{A \mid S} \phi_{B \cup S}$, and $\psi = \phi_{A \mid S} \phi_{B \mid S} \phi_S$, respectively. The relevant marginal and conditional graphs are $a \to e$ for $S$ and the ones of Figure 8. Forming the independence graph of $\psi$ through graph union, we find the three possible solutions displayed in Figure 5a–c corresponding, respectively, to combination alternatives (1)–(3) with the modifications that $a - d$ (solutions (a) and (c)) and $a - b$ (solutions (b) and (c)) have been replaced with $a \to d$ and $a \to b$ to avoid directed cycles. (Note that these modifications do not alter the represented independence statements.) Since we shall prefer solutions representing the largest sets of independence statements, there is a clear preference order among the three alternatives (solution (a) is preferable to solution (b) which is preferable to solution (c)).

A similar analysis can be performed for the dyspnoea example in the Introduction. Again there appears to be a clear preference order among the solutions, with the optimal solution displayed in Figure 1b.

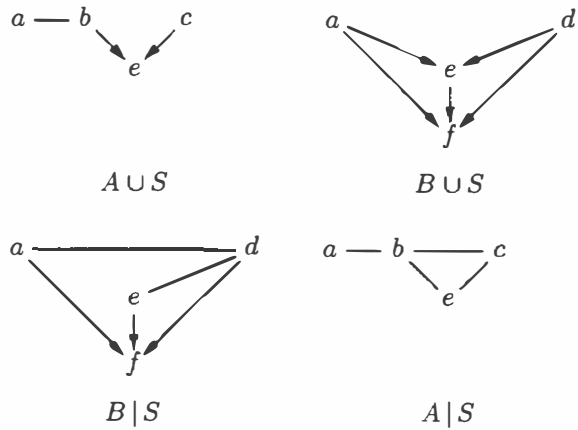

$A \cup S$    $B \cup S$

$B \mid S$    $A \mid S$

Figure 8: Marginal and conditional independence graphs of the graph of Figure 3a with $A = \{a, b, c, e\}$, $B = \{a, d, e, f\}$, and $S = \{a, e\}$.

### 3.6 COMPONENT POTENTIALS

Given a joint belief, $\psi$, and a chain graph, $\mathcal{G}$, obtained through enforcement of one or more conditional independence assumptions, we wish to determine an associated set of component potentials. Furthermore, we have available a set of belief potentials associated with a junction tree corresponding to $\mathcal{G}$.

Notice that if $\psi$ and $\mathcal{G}$ are produced as described in Sections 3.3–3.5, $\psi$ is guaranteed to factorize according to $\mathcal{G}$. That is, there exist component potentials $\xi_A$ such that $\psi \propto \prod_A \xi_A$ (cf. Equation (1)).

Following Equation (3) the problem can be divided into $n$ subproblems, where $n$ is the number of chain components of $\mathcal{G}$. More specifically, since $\psi = \prod \psi(K \mid \mathrm{pa}(K))$, we must determine potentials $\xi_A$ for each chain component $K$ such that

$$\psi_{K^+} = \psi_{\mathrm{pa}(K)} \left( \prod_{A \in \mathcal{A}_K} \xi_A \right) \Big/ \left( \sum_K \prod_{A \in \mathcal{A}_K} \xi_A \right)$$

(cf. Equation (4)), where $K^+ = K \cup \mathrm{pa}(K)$ and $\mathcal{A}_K$ is the set of cliques in $(\mathcal{G}_{K^+})^m$ containing at least one node in $K$. Notice that, since belief potentials are available, $\psi_{K^+}$ can be computed.

The potentials $\xi_A$ can be found via Möbius inversion when $\psi_{K^+}$ is positive; see e.g. Lauritzen & Wermuth (1989). Unfortunately, this is rarely the case. However, it seems plausible that an extended version of the Möbius inversion exists when $\psi_{K^+}$ is known to factorize according to $\mathcal{G}$.

Lauritzen & Wermuth (1984) has shown that for any decomposable graphical chain model there exists an equivalent recursive model; that is, if $(\mathcal{G}_{K^+})^m$ is triangulated for each $K$. Thus, if $\mathcal{G}$ is decomposable, we may generate the equivalent DAG and compute conditional probabilities (component potentials) $\psi(v \mid \mathrm{pa}(v))$ (cf. Equation (2)). If $\mathcal{G}$ is not decomposable, we may triangulate each subgraph $(\mathcal{G}_{K^+})^m$ by inserting fill-ins and then generate a DAG, $\mathcal{G}^*$, from the resulting graph. In the latter case the resulting recursive model will be suboptimal in two ways. First, $\mathcal{G}^*$ fails to represent all the independence statements represented by $\mathcal{G}$. Second, the computational complexity imposed by the optimal triangulation of $(\mathcal{G}^*)^m$ is at least as large as the computational complexity imposed by the optimal triangulation of $\mathcal{G}^m$, since the triangulation of each $(\mathcal{G}_{K+})^m$ 'constrains' the triangulation.

## 4 EXPERIMENTS

Since, from a theoretical point of view, not much can be said about the practical importance of link removal, we shall now report on some results of an empirical study conducted on a number of real-world networks.

The networks are Pathfinder (Heckerman, Horvitz & Nathwani 1992) (including 109 nodes) for diagnosing lymph node pathology, two subnetworks of MUNIN (Andreassen, Woldbye, Falck & Andersen 1987) (including about 190 nodes each) for diagnosing disorders in the peripheral nervous system, and a time-sliced network model of the biological processes of a water treatment plant including 32 process variables (Jensen, Kjærulff, Olesen & Pedersen 1989).

The criterion applied for selecting a link $\alpha \sim \beta$ to be removed from a clique $C$ is based on the reduction of the total state space and the 'distance' between the exact, $\phi_C$, and the approximate, $\psi_C$, clique potentials. The distance, $D(\phi_C, \psi_C)$, is measured as the conditional mutual information between $\alpha$ and $\beta$ given $C \setminus \{\alpha, \beta\}$ (also: the Kullback-Leibler divergence between $\phi_C$ and $\psi_C$) given as

$$I(\alpha, \beta \mid C \setminus \{\alpha, \beta\}) = D(\phi_C, \psi_C) = E \log(\phi_C/\psi_C)$$

with expectation taken with respect to $\phi$, and where $I = 0$ when $\phi_C = \psi_C = 0$. A useful relationship between the absolute divergence and the Kullback-Leibler divergence (see e.g. Kullback (1967)) states that

$$|\phi_A - \psi_A| \leq \sqrt{\tfrac{1}{2} D(\phi_C, \psi_C)} \qquad (8)$$

for any $A \subseteq C$.

In the experiments links with lower mutual information were preferred, and savings (reduction of state space) were used only to break ties. Further, links were removed until a total divergence of at most 0.001 was reached (the total divergence after a series of removals equals the sum of the individual divergences (Kjærulff 1993)). Using Inequality (8) the theoretically upper bound on the absolute error is found to be 2%.

Table 1 displays the results. The 'size' of a network equals the sum of the sizes of the state spaces of the cliques after sensible triangulation. For all networks except MUNIN2 the Kullback-Leibler divergences were computed using exact clique potentials.

For the MUNIN2 network simulated potentials based on 10,000 iterations of forward sampling were used.

| Network (size) | Links removed | Reduction |
|---|---|---|
| Pathfinder (187, 244) | 26 | 36.4% |
| MUNIN1 (2, 302, 119) | 145 | 34.3% |
| MUNIN2 (183, 549, 219) | 190 | 96.0% |
| Water (9, 443, 571) | 126 | 97.2% |

Table 1: Empirical results of applying link removal to real-world networks.

The savings obtained for the Pathfinder and the MUNIN1 networks are relatively modest, whereas significant savings are obtained for the MUNIN2 and the Water networks. The reduction from 183.5 M to 7.3 M for the MUNIN2 network makes it possible to perform exact computations using the junction-tree methodology. The large savings for the MUNIN2 and the Water networks are due partly to the fact that a number of the orphan nodes are instantiated to their 'normal' state.

## 5 DISCUSSION

An important feature of a clique-potential approximation is the attenuation of its impact with increasing distance from the target clique (Kjærulff 1993). This feature is especially important in connection with time-sliced Bayesian networks. An additional property of the method, is the property of errors remaining localized in absence of posterior evidence and, under certain conditions, even in presence of posterior evidence (Kjærulff 1993).

The presented approximation method has been compared with the method suggested by Jensen & Andersen (1990). Briefly, their method is based on annihilation of small probabilities by setting the $k$ smallest probabilities to zero for each clique potential of a junction tree, where $k$ is chosen such that the sum of the $k$ smallest probabilities is less than a predetermined threshold. After annihilation, the belief tables are compressed in order to take advantage of the introduced zeros.

The comparison (reported in Kjærulff (1993)) demonstrates that link removal in some cases is significantly better than annihilation. In other cases, however, a comparison turns out to the disadvantage of link removal. Intuitively, this seems absolutely reasonable, since a model including links representing weak dependences will be almost equivalent to a model which lacks these links, but it might be quite different from a model obtained by uniformly removing a corresponding amount of probability mass from the belief tables. On the other hand, link removal is unsuited in cases where there are no 'weak links'. Thus, to approximate a given network using these two methods, link removal should be tried first and when all 'weak links' have been removed, annihilation should take over.

Application of link removal does not require the construction of exact clique potentials (as opposed to annihilation). Further, the creation of simulated clique potentials (through e.g. forward sampling) and possible subsequent link removal provides a way of establishing an annihilated and compressed junction tree representation of a network without first creating exact potentials.

Inequality (8) is essential, since the key indicator associated with an approximation is most often the maximum absolute error. However, under arrival of posterior evidence, the inequality can only be used as a rough guideline. Thus, among directions for future research, an important one is assessment of a good upper bound on the error given evidence.

### Acknowledgements

I am indebted to Steffen L. Lauritzen for providing many valuable comments on earlier drafts. The research has been funded partly by the Danish Research Councils through the PIFT programme.